\documentclass[letterpaper]{article} 
\usepackage{AAAI2024/aaai24}  
\usepackage{times}  
\usepackage{helvet}  
\usepackage{courier}  
\usepackage[hyphens]{url}  
\usepackage{graphicx} 
\def\UrlFont{\rm}  
\usepackage{natbib}  
\usepackage{caption} 
\pdfoutput=1

\usepackage{times}
\usepackage{latexsym}

\usepackage[T1]{fontenc}

\usepackage[utf8]{inputenc}

\usepackage{microtype}
\usepackage{inconsolata}

\usepackage{textcomp}
\usepackage[utf8]{inputenc}
\usepackage{amsmath,amssymb}
\usepackage[ruled]{algorithm2e}
\usepackage{algorithmic}
\usepackage{graphicx}
\usepackage{array}
\usepackage{booktabs}
\usepackage{float}
\usepackage{caption}
\usepackage{natbib}
\usepackage{cite}
\usepackage{multirow}
\usepackage{comment}
\usepackage{mathtools}

\usepackage{amsfonts}
\usepackage{url}

\usepackage{tabularx}

\DeclareMathOperator*{\argmax}{\mathrm{argmax}}

\usepackage{url} 
\usepackage{color}

\newcommand{\tabincell}[2]{\begin{tabular}{@{}#1@{}}#2\end{tabular}}
\usepackage{newfloat}
\usepackage{listings}
\DeclareCaptionStyle{ruled}{labelfont=normalfont,labelsep=colon,strut=off} 
\floatstyle{ruled}
\newfloat{listing}{tb}{lst}{}
\floatname{listing}{Listing}
\title{AAAI Press Anonymous Submission\\Instructions for Authors Using \LaTeX{}}
\author{
    Renlong Jie \textsuperscript{\rm 1, \rm 2}\footnotemark[1],
    Xiaojun Meng \textsuperscript{\rm 1}\footnotemark[2],
    Xin Jiang \textsuperscript{\rm 1},
    Qun Liu \textsuperscript{\rm 1}
}

\affiliations{
    \textsuperscript{\rm 1}Huawei Noah's Ark Lab\\
    \textsuperscript{\rm 2}Northwestern Polytechnical University, China\\

    jrl@nwpu.edu.cn\\
    \{xiaojun.meng, Jiang.Xin, qun.liu\}@huawei.com
}

\usepackage{url} 
\usepackage{color}

\title{Unsupervised Extractive Summarization with Learnable Length Control Strategies}

\begin{document}

\newcolumntype{L}[1]{>{\raggedright\arraybackslash}p{#1}}
\newcolumntype{C}[1]{>{\centering\arraybackslash}p{#1}}
\newcolumntype{R}[1]{>{\raggedleft\arraybackslash}p{#1}}
\maketitle

\renewcommand{\thefootnote}{\fnsymbol{footnote}} 
\footnotetext[1]{Work is done as a postdoctoral fellow of Noah's Ark Lab, Huawei.}
\footnotetext[2]{Corresponding authors.}
\renewcommand{\thefootnote}{\arabic{footnote}} 

\begin{abstract}

Unsupervised extractive summarization is an important technique in information extraction and retrieval. Compared with supervised method, it does not require high-quality human-labelled summaries for training and thus can be easily applied for documents with different types, domains or languages. Most of existing unsupervised methods including TextRank and PACSUM rely on graph-based ranking on sentence centrality. However, this scorer can not be directly applied in end-to-end training, and the positional-related prior assumption is often needed for achieving good summaries. In addition, less attention is paid to length-controllable extractor, where users can decide to summarize texts under particular length constraint. This paper introduces an unsupervised extractive summarization model based on a siamese network, for which we develop a trainable bidirectional prediction objective between the selected summary and the original document. Different from the centrality-based ranking methods, our extractive scorer can be trained in an end-to-end manner, with no other requirement of positional assumption. In addition, we introduce a differentiable length control module by approximating 0-1 knapsack solver for end-to-end length-controllable extracting. Experiments show that our unsupervised method largely outperforms the centrality-based baseline using a same sentence encoder. In terms of length control ability, via our trainable knapsack module, the performance consistently outperforms the strong baseline without utilizing end-to-end training. Human evaluation further evidences that our method performs the best among baselines in terms of relevance and consistency.

\end{abstract}

\section{Introduction}

Text summarization aims at producing a short and clear text that expresses the major ideas and important information of an original article. There are two types of summarization methods: 1) \textit{abstractive summarization} directly generates summary texts as a sequence,
while 2) \textit{extractive summarization} extracts the most important sentences from the original article to form a summary. As known to all, recent abstractive models like GPT-style large language models have shown an incredible ability in text generation~\citep{zhang2023benchmarking}. These LLMs can be used to summarize everything with designing certain prompts.
However, it is noticed that compared with abstractive summarization with LLMs, extractive models are in much smaller size and thus have the advantages of faster inference and less demand on computing costs. In additional, factually consistent summaries can not be guaranteed by LLMs~\citep{tam-etal-2023-evaluating}.
Oppositely, extractive models can ensure factual consistency to the original article and the extracted text content is more controllable for the safe use of real products.

Extractive summarization includes traditional unsupervised methods like TextRank ~\citep{mihalcea2004graph, erkan2004lexrank}, and supervised methods like BertSum~\citep{liu2019fine}. 
Although being capable of building high-quality summaries, supervised methods rely on high-quality human-labelled datasets. However, it is not realistic to label these datasets across all kinds of domains, document types and user requirements. 
The mainstream existing unsupervised summarization methods are centrality-based methods, including Textrank and PACSUM \citep{mihalcea2004textrank, zheng2019sentence}. They are no-differentiable methods and thus can not be trained with end-to-end manner. In addition, some of them like PACSUM and FAR \citep{zheng2019sentence, liang2021improving} rely on the positional-related prior assumption to achieve good performance. 

Building upon existing summarization work, \textit{length-controllable summarization}~(\textbf{LCS})~\citep{yu2021lenatten} has been studied to control the length of output summaries. The LCS approach enables users to control summary length, thus to serve various real industrial scenarios when users read different types of articles, or use different size of screens, on different occasions. However, very few work~\citep{mehdad2016extractive} targets at the LCS problem of extractive methods, and it is hard to annotate summaries under strict length control. In fact, the nature of extracting summary texts under length constraint is more like a combinatorial optimization problem such as \textit{Knapsack Problem}~\citep{martello1990knapsack}, where we extract the most valuable sentences within the length constraint. This is traditional solved by dynamic programming (DP), but this solver can not be applied as a differentiable module in end-to-end deep learning architectures.


In this paper, we aim at developing a novel unsupervised extractive summarization method with strong length control ability. We propose an unsupervised summarization architecture based on Simple Siamese (SimSiam)~\citep{chen2021exploring}, and develop a bidirectional prediction target to replace the traditional centrality target. Meanwhile, we model the LCS of extractive methods by a 0-1 Knapsack Problem, and introduce a transformer encoder to approximate the dynamic programming solver for end-to-end training. The main contributions of this paper are listed as follows:



\begin{itemize}
\setlength{\itemsep}{0pt}
\setlength{\parsep}{0pt}
\setlength{\parskip}{0pt}

\item We propose an end-to-end trainable unsupervised model for extractive summarization, via a transformer-based siamese network. A bidirectional representation prediction objective is introduced for model training, which is more flexible than the traditional centrality by simply summing up the similarities with all other sentences. 
entences scores give\item We also introduce a novel transformer model to approximate the knapsack solver with high accuracy, which can be applied as a differentiable module in various deep architectures. By jointed training with the proposed extractive summarization model, the length control strategy can be learned in a differentiable way. 
\item Experiments show that our proposed unsupervised method can produce high-quality summaries comparable to the state-of-the-art unsupervised methods in automatic evaluation. 
With the trainable knapsack module, our model performance of length control outperforms related baselines without utilizing end-to-end training, and even outperforms the supervised method in human evaluation.
\end{itemize}

\section{Related Work}\label{Sec:2}


\subsection{Extractive Summarization}
Extractive summarization \citep{erkan2004lexrank,liu2019fine,zhong2020extractive} output text summaries by extracting the most informative segments from the original text. Based on the requirement of labelled data, it can be classified into unsupervised methods and supervised methods. For unsupervised methods, centrality is widely used as a metric for sentence extraction \citep{mihalcea2004textrank, zheng2019sentence, liang2021improving}. Traditional centrality based method such as TextRank applies similarity-based centrality to determine the importance of each sentence, but it usually can not provide good performance when evaluated with human labelled ground truth. \citet{zheng2019sentence} introduced the prior assumption that the contribution of two nodes' connection to their respective centrality is influenced by their relative position, which lead to a large improvement of the automatic metrics on extractive summarization. \citet{liang2021improving} introduced a facet-aware mechanism to handle the case where the document has multiple facets. \citet{xu2020unsupervised} applies a pretrained hierarchical transformer to calculate the probability of containing each sentence in the context, as well as attention based centrality as criteria for scoring sentences. Optimal transport is also applied in as a non-learning-based method \citep{tang2022otextsum}. For supervised methods, bidirectional sequence models like bi-LSTM or transformer encoder can be applied for encoding and scoring the representation of each sentence in the document \citep{liu2019fine}. Graph-based methods and reinforcement learning have been introduced for further improvement \citep{hetergraph2020, multiplexgraph2021, gu2022memsum}.

\subsection{Length Control in Summarization}
The majority of work in LCS focus on abstractive methods, which aims to stop the generating process at desired length by generating a stop token (\textit{e.g.,} \texttt{[EOS]}) \citep{rush-etal-2015-neural, kikuchi-etal-2016-controlling, liu2018controlling, takase2019positional, yu2021lenatten, makino-etal-2019-global, liu2022length}. 
Previous studies on length-controllable extractive methods are very limited. \citet{mehdad2016extractive} shows a comparison of various techniques for extractive summarization under the strict length constraint. They propose a supervised learning method to score each sentence using a combination of pre-computed features, and extract sentences into a final summary in a greedy fashion based on their scores while respecting the length constraint.

\section{Methodology}\label{Sec:3}


\begin{figure*}[htb]
\begin{center}
 \includegraphics[width=0.8\linewidth]{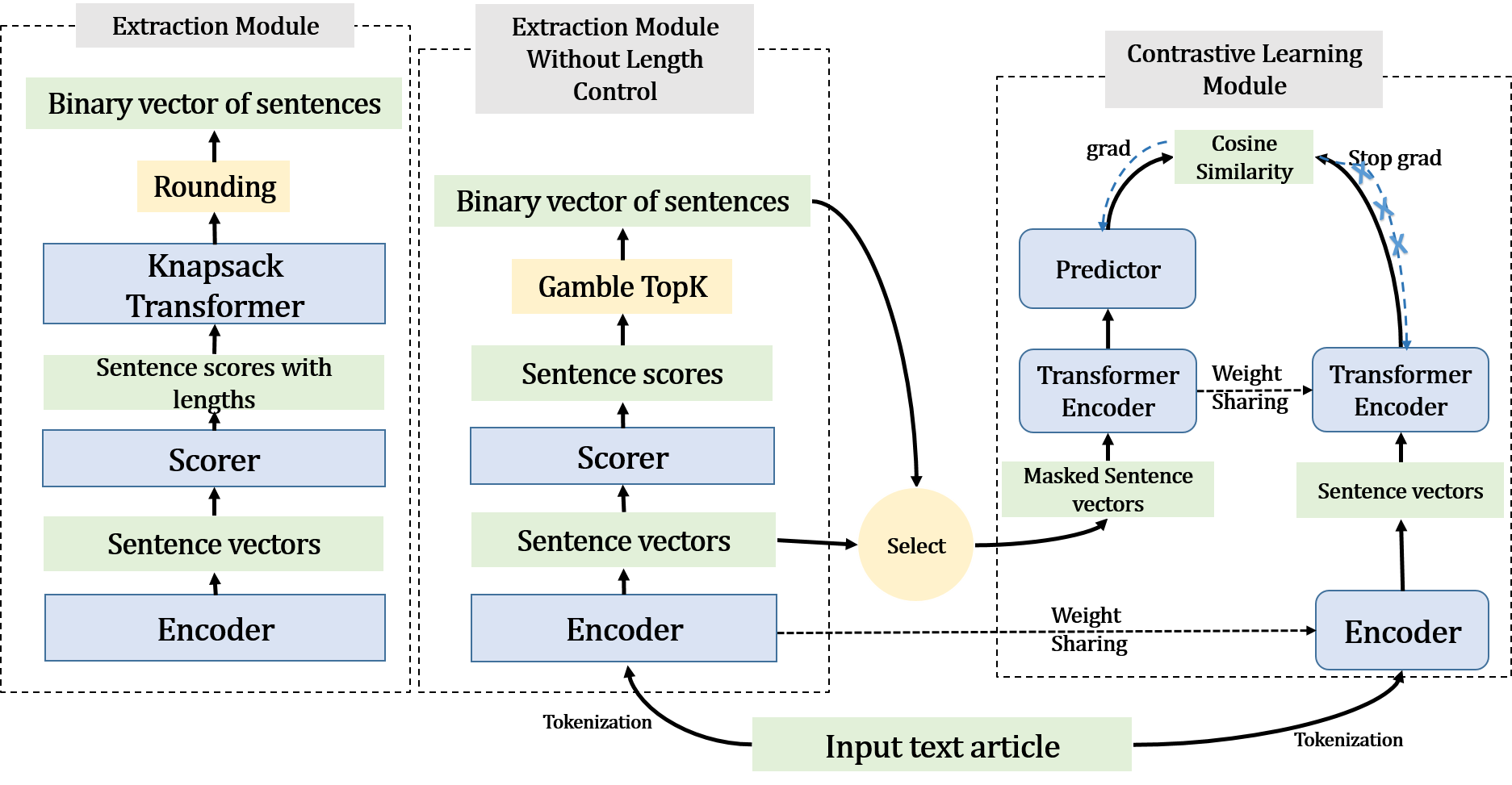}
 \caption{The diagram of the joint siamese model for unsupervised extractive summarization with and without length constraint. For the case with length control, the Knapsack transformer is applied in training time, which takes sentence scores and sentence lengths as input, and outputs the indices of extracted sentences. The unsupervised learning is to train the scorer by maximizing the prediction accuracy of SimSiam-based contrastive learning module with two proposed objectives (\textbf{Rest-Ext} and \textbf{Ext-Doc}). 
 } \label{Fig:2}
\end{center}
 \vspace{-1.5em}
\end{figure*}

\subsection{Problem formulation} \label{Sec:3.1} 

Consider a document $D$ containing a set of $n$ sentences $\{x_1, ..., x_n\}$ with lengths $\{l_1,...,l_n\}$. The basic extractive summarization is to select a subset of sentences to formulate a summary 
$S = \{x_i\}_{i\in I_{ext}}$, which is an ordered set of selected sentences in $D$ given by indices $I_{ext} \subseteq \{1,...,n\}$. Define $f$ as the evaluation function, $V$ is the optimal subset of extracted summary, for general unsupervised extraction \textbf{without length control}, the learning objective is given by: 
\begin{equation}
\setlength{\abovedisplayskip}{3pt}
\setlength{\belowdisplayskip}{3pt}
\begin{split}
    V = \argmax_{S \subseteq D}f(S),\,\text{s.t. }n(S) \leq M,
\end{split}
\label{Eq:1}
\end{equation}
where $M$ is the maximum number of sentences to be extracted. For the summarization \textbf{with the length control} as a requirement, the total length of the extracted summary should not exceed a pre-determined budget $C$. Thus, Eq.\eqref{Eq:1} should be subject to $\sum_{i\in S} l_i < C$. Existing extractive approaches such as LexRank~\citep{erkan2004lexrank} and BertSum~\citep{liu2019fine} assign a score to each sentence based on measuring the importance of it in the whole document. 
Assume that the goodness of the summary is represented by the sum of scores of selected sentences, the evaluation function $f$ can be re-written as $f(S) = \sum_{i\in S}v_i$, and the learning objective with length control becomes: 
\begin{equation}
V = \argmax_{S \subseteq D}\sum_{i\in S}v_i,\,\text{s.t. }\sum_{i\in S} l_i < C,
\end{equation}
where $v_i$ is the score of the $i$-th sentence given by a scorer. By now, this length controllable extractive summarization emerges as the classical 0-1 Knapsack Problem.

\subsection{Unsupervised Method} \label{Sec:3.4}
For the traditional graph-based ranking method on extractive summarization, the degree of centrality for a sentence $i$ is calculated by a simple summation over all its similarities with other sentences as follows: 
\begin{equation}
\textbf{centrality}(x_i) = \sum_{j\in \{1,...,i-1,i+1,...n\}}e_{ij},
\end{equation}
where $e_{ij}$ is represents the similarity between sentence $i$ and sentence $j$. This text centrality is further used to measure the importance of each sentence.

\paragraph{Motivation.} We notice that the similarity $e_{ij}$ here can also be Pearson correlation between each pair of sentence representations, which further indicates that the sentences with higher centrality can be explained or even predicted by the whole set of other sentences with a relatively high possibility. Meanwhile, the representation of the extracted summary should have a strong correlation with the representation of the whole document by the nature of summaries. Motivated by this, we propose a bidirectional prediction criterion to replace the traditional centrality in unsupervised summarization. 
The first prediction criterion (\textbf{Rest-Ext}) is that each of the selected sentences for extractive summary, rather than unselected sentences, can be better predicted by the rest of sentences (selected or unselected) in the document. The second prediction criterion (\textbf{Ext-Doc}) is that the whole set of the selected sentences can better predict the representation of the whole document compared to other unselected ones. Overall, with these two objectives, the learning process is to train the sentence scorer and predictor to increase the prediction accuracy, thus extracting the most importance sentences to form as a summary in an unsupervised manner, as in the Figure~\ref{Fig:2}.

Particularly, consider the final sentence representations of the original document is $\mathbf{H}^S = (\mathbf{h}^S_1, ..., \mathbf{h}^S_N)$, the rest of the document by excluding one randomly sampled sentence from the extracted summary is $\mathbf{H}^{S'} = (\mathbf{h}^{S'}_1, ..., \mathbf{h}^{S'}_N)$, and the extracted summary is $\mathbf{H}^{S''} = (\mathbf{h}^{S''}_1, ..., \mathbf{h}^{S''}_N)$, where we preserve the size of representations $N$ but mask out unrelated information for each case in a lower representation level. 
We introduce two learning objectives: (a) The first predictive loss (\textbf{Rest-Ext}) of using the final representation of the rest of sentences $\mathbf{H}^{S'}$ to predict the representation of a randomly sampled one from selected sentences. 
\begin{equation}
\mathbf{\hat{h}}^S_i = \text{Predictor}(\mathbf{h}^{S'}_1,...,\mathbf{h}^{S'}_N),
\label{Eq:pred_1}
\end{equation}
where $i$ is the index sampled from the selected sentences $i\in I_{ext}$. Here we apply the random sampling as we assume each of the selected sentences can be well predicted given other sentences of this document. We select the $i$-th top vector of the predictor as the predicted representation, where the predictor is a transformer encoder. Meanwhile, to enlarge the relative advantage of selected sentences, and avoid only learning the predictor rather than this selecting mechanism, we also consider a prediction of negative sample $\mathbf{\hat{h}}^S_{i'}$ with the same masking mechanism, where $i'$ is the index of a randomly sampled unselected sentence with low scores. Thus, the predictive loss is to be:
\begin{equation}
L_{Rest-Ext} = -\text{cos}(\mathbf{\hat{h}}^S_i, \mathbf{h}^S_i) + |\text{cos}(\mathbf{\hat{h}}^S_{i'}, \mathbf{h}^S_i)|,
\end{equation}
where $\mathbf{\hat{h}}^S_i$ is the $i$-th top sentence vector of the predictor from the extraction side (the left part of contrastive learning module in the Figure~\ref{Fig:2}) and $\mathbf{h}^S_i$ is the $i$-th top sentence vector of the sentence-level encoder from the whole document context-aware side (similarly, the right part of that contrastive learning module). 
The cosine similarity is given by $cos(\mathbf{a},\mathbf{b})=\mathbf{a}
\cdot~\mathbf{b}/||\mathbf{a}||\cdot||\mathbf{b}||$ for any pair of vector $\mathbf{a}$ and $\mathbf{b}$. Note that minimizing $L_{Rest-Ext}$ is equivalent to optimize a MSE loss between normalized vector $\mathbf{a}$ and $\mathbf{b}$ as $||\mathbf{a}-\mathbf{b}||^2_2 = 2-2 \mathbf{a}\cdot\mathbf{b}/(||\mathbf{a}||_2\cdot||\mathbf{b}||_2)=2-2cos(\mathbf{a}, \mathbf{b})$. We use the absolute value of prediction similarity from negative samples to achieve low absolute correlation.

(b) The second predictive loss (\textbf{Ext-Doc}) of using all the representations of selected sentences to predict the document representation. In our study, we apply the average of sentence representations as the document representation, thus we have:
\begin{equation}
\begin{split}
\mathbf{\hat{h}^D} = \frac{1}{N}\sum_i^M\mathbf{\hat{h}_i^S} = \text{Predictor}(\mathbf{h}^{S''}_{1},...,\mathbf{h}^{S''}_{N})
\end{split}
      \label{Eq:pred_2}
\end{equation}
as a prediction. The corresponding loss is:
\begin{equation}
L_{Ext-Doc} = -\text{cos}(\mathbf{\hat{h}^D}, \mathbf{\bar{h}}^S) + |\text{cos}(\mathbf{\hat{h'}^D},\mathbf{\bar{h}}^S)|,
\end{equation}
where $\mathbf{\bar{h}}^S = \frac{1}{N}\sum_{i=1}^M(\mathbf{h}^S_i)$ is the average sentence representation of the full document, and $\mathbf{\hat{h'}^D}$ is the predicted representation given by the set of unselected sentence representations. Overall, the total loss of training objective is:
\begin{equation}
L = \lambda_1 L_{Rest-Ext} + \lambda_2 L_{Ext-Doc},
\label{Eq:final_loss}
\end{equation}
where $\lambda_1$ and $\lambda_2$ are two hyper-parameters. 


To get the final representations $\mathbf{H}^S$, $\mathbf{H}^{S'}$ and $\mathbf{H}^{S''}$ for bidirectional prediction with a trainable selector, we develop a SimSiam architecture as shown in the Figure~\ref{Fig:2}. In general, it consists of a sentence extraction module, and a contrastive learning module. Sentence extraction module takes a tokenized document $\{x_1, ..., x_n\}$ as the input, in which a sentence-level encoder such as BERT or SimCSE processes sentences and outputs the first-level sentence representation $\mathbf{E} = (\mathbf{e}_1,...,\mathbf{e}_n)$ by taking the top vectors of \texttt{[CLS]} token. A transformer-based scorer provides importance scores for each sentence. These scores are used for selecting the sentence representations in a differentiable manner, which will be discussed later. The selected sentence vectors $\mathbf{E''}$, sentence vectors excluding a randomly sampled one $\mathbf{E'}$ and all sentence vectors $\mathbf{E}$ are passed through another transformer-based encoder in the contrastive learning module, to get the second-level sentence representations $\mathbf{H}^{S''}$, $\mathbf{H}^{S'}$ and $\mathbf{H}^S$ for later cosine similarity. Note that our so-called first-level and second-level sentence representations are separated, as they serve different purposes in the training. The first-level one is used for scoring and selecting sentences, while the second-level is mainly learned to serve our designed bidirectional prediction criterion.
The \texttt{``Weight}~\texttt{Sharing''} in Figure~\ref{Fig:2} denotes that the weight in each group of encoders is shared. In inference, only the encoder and scorer is needed.

\subsection{Trainable Length Controller}\label{Sec:3.2}


As discussed in Problem Formulation, length controller extractive summarization can be modelled by the Knapsack Problem~\citep{martello1990knapsack}. The classical Knapsack Problem is often solved by adopting the dynamic programming algorithm, however which is non-differentiable. Recent studies have explored neural knapsack models \citep{nomer2020neural, hertrich2021provably, li2021novel}. 
\citet{xu2020deep} use NNs to enhance dynamic programming and 
\citet{hertrich2021provably} present a detailed mathematical study to investigate the expressivity of NNs, and find that a class of RNNs with feedforward RELU compute provably good solutions to the NP-hard Knapsack Problem. In addition, \citet{yildiz2022reinforcement} uses reinforcement learning to achieve near-optimum solutions in a much faster way.
Previous work of neural knapsack motivates us to rethink the LCS problem and address it with a trainable end-to-end neural network. 

In this paper, 
we introduce a novel knapsack transformer to approximate the dynamic programming solver.
We apply BERT-like transformer encoder~\citep{vaswani2017attention, devlin2018bert} as the approximator, and simulate data for training it.
\begin{figure}[th]
\begin{center}
 \includegraphics[width=0.7\linewidth]{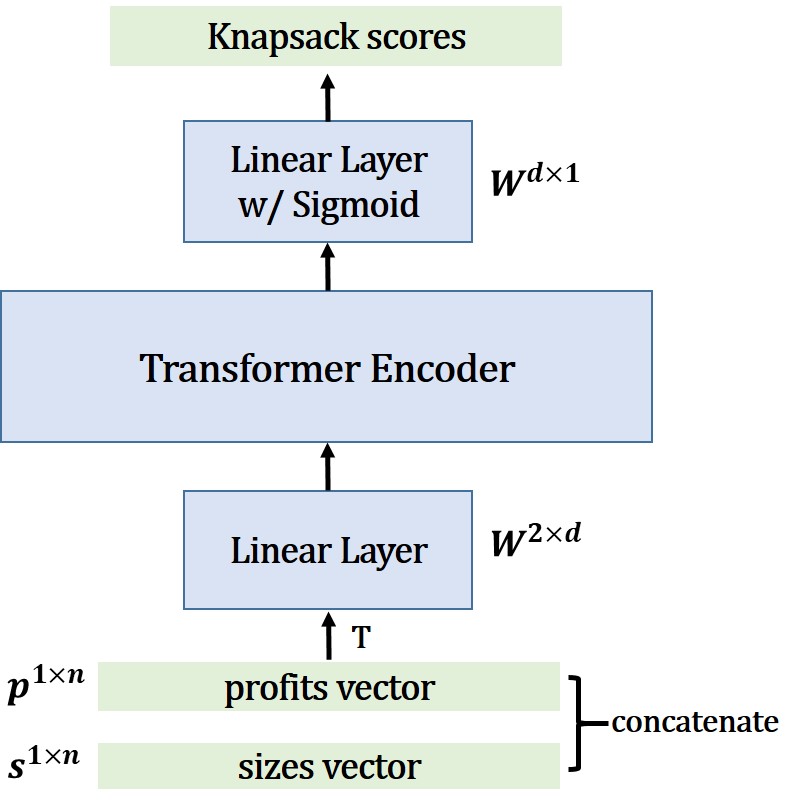}
 \caption{The proposed knapsack transformer. A linear layer is added to convert the feature dimension of the concatenated input to the model dimension. In our LCS problem, the profit vector represents each sentence score and the size vector is its length. ``T'' denotes to transpose the matrix.} \label{Fig:1}
\end{center}
 \vspace{-1.5em}
\end{figure}
The structure of knapsack transformer is presented in Figure~\ref{Fig:1}.
For each training sample, the profits and sizes of items are concatenated into a 2 by $n$ matrix. Prior to the original transformer encoder, a linear layer is applied to map the input feature dimension of 2 to the hidden embedding dimension $d$. After the transformer module, another linear layer is applied to map the hidden states from dimension $d$ back to single dimension, followed by a Sigmoid activation. By now, the knapsack score of each sentence is outputted. The training objective is to minimize a binary cross entropy loss between the model output and the binary vector of the ground truth. To make the trained model generalize well, the profits for each sample are normalized and sum up to 1, while the sizes are normalized by dividing the target total size.
In the inference, these knapsack scores are rounded to 0 or 1 corresponding to not select or select the item.

\subsection{Enabling Differentiable End-to-End Training}\label{Sec:3.5}

Figure~\ref{Fig:2} demonstrates our joint SimSiam model for unsupervised extractive summarization and the extractive module with or without length-controllable ability (\textit{i.e.,} using a pre-trained knapsack transformer). 
We apply gumble softmax and straight-through trick to enable the end-to-end training with a differentiable process of sentence representation selection. For unsupervised model \textbf{without knapsack transformer}, we use topK sampling to get the indices of selected sentences in inference. However, in training, Gumbel-soft topK sampling is applied for better robustness~\citep{jang2017categorical, kool2019stochastic} as follows:
\begin{equation*}
\setlength{\abovedisplayskip}{3pt}
\setlength{\belowdisplayskip}{3pt}
    \begin{split}
        &y_i = \text{GumbleSoftmax}(\pi_i, T),\,i=1,2,...,M\\
        &\mathbf{s}_{tk} = \text{TopK}(y),
        \end{split}
\end{equation*}
where $\pi_i$ is the score of sentence $i$ given by BertSum module, $\mathbf{s}_{tk}$ is the binary vector showing the position of topK indices in $\mathbf{y}$. Then we can use a straight through trick given by:
\begin{equation*}
\setlength{\abovedisplayskip}{3pt}
\setlength{\belowdisplayskip}{3pt}
    \hat{\mathbf{s}} = (\mathbf{s}_{tk} - \mathbf{y})_f + \mathbf{y},
\end{equation*}
to get a derivable binary vector $\hat{\mathbf{s}}$, where $(.)_f$ means we do not calculate gradients within the parentheses in backward propagation. In forward propagation, modules using this topK can only see a binary vector $\hat{\mathbf{s}} \in \mathbb{B}^n$ containing a binary value per sentence, which means to select (\textit{i.e.,} rounding to 1) or not select (\textit{i.e.,} rounding to 0) this sentence to form the resulting summary. In backward propagation, the gradients are calculated with the continuously differentiate value of $\mathbf{y}$. 

For our model \textbf{with knapsack transformer} shown in Figure~\ref{Fig:1},
the concatenated input of score vector and length vector is fed into the pre-trained knapsack transformer for selecting sentences as the summary. Note that there is a step-wise rounding operation on the output score of knapsack transformer, where we similarly apply the straight through trick to enable back propagation:
\begin{equation*}
\setlength{\abovedisplayskip}{3pt}
\setlength{\belowdisplayskip}{3pt}
    \hat{\mathbf{s}} = (\mathbf{s}_{\text{round}} - \mathbf{s})_f +  \mathbf{s},
\end{equation*}
where $\mathbf{s}$ is the output score vector of the knapsack transformer, and $\mathbf{s}_{\text{round}}$ is its rounded vector with elements as either 0 or 1. Similar to the above-mentioned topK, the $\hat{\mathbf{s}}$ is differentiable given that the $\mathbf{s}$ is differentiable. To optimize $L_{Ext-Doc}$, the binary vector $\hat{\mathbf{s}}$ representing the selected sentences is multiplied with the representation of that sentence with a row-wise scalar multiplications by the following masking: 
\begin{equation}
\setlength{\abovedisplayskip}{3pt}
\setlength{\belowdisplayskip}{3pt}
    \mathbf{e}''_i = \hat{s}_i \mathbf{e}_i,\quad \hat{s}_i \in \{0,1\},\, i\in \{1,...,n\},
\end{equation}
where $\hat{\mathbf{s}}_i$, $\mathbf{e}_i$ and $\mathbf{e}''_i$ are the $i$-th element in $\hat{s}$, the vector for $i$-th sentence in $\mathbf{E}$, and the new masked vector for $i$-th sentence in $\mathbf{E''}$ after multiplication. It means that only the embedding of selected sentences are kept. Then the vectors of extracted summary $\mathbf{E''}$ are passed to the contrastive learning module.
To optimize $L_{Rest-Ext}$, each time we mask the vector of a randomly sampled selected sentence indexed by $k$ to get the rest representations for the predictor through another row-wise multiplication: 
\begin{equation}
\setlength{\abovedisplayskip}{3pt}
\setlength{\belowdisplayskip}{3pt}
    \mathbf{e}'_i = (1-\tilde{s}_i) \mathbf{e}_i,\quad \tilde{s}_i \in \{0,1\},\, i\in \{1,...,n\},
\end{equation}
where $\tilde{s}_i = (\delta_{ik}-s_i)_f + s_i$, and $k\in I_{ext}$ is a single index sampled from the set of selected sentence indices by Gumble-topK or rounding. 
$\delta_{ik}$ equals to $1$ only when $i=k$ otherwise it equals to $0$.
Thus $\mathbf{e}'_i$ is the new masked vector for the $i$-th sentence in $\mathbf{E}'$, where the $k$-th representation is masked by a 0 vector. Finally, we are able to get the input representations for two training objectives in contrastive learning, and the score-based selecting process is fully differentiable.

\section{Experiment}\label{Sec:4}


We conduct experiments on CNNDM dataset \citep{hermann2015teaching}, NYT dateset \citep{durrett2016learning} and CNewSum dataset \citep{wang2021cnewsum} for news summarization in both English and Chinese. 

\subsection{Training of Knapsack Network}\label{Sec:4.2.1}

We simulate 6 million training samples including profits and sizes of each item according to the empirical distribution of lengths and scores in the context of extractive summarization. 
As there is a large margin between average sentence lengths of CNewSum and CNNDM/NYT, we separately create training data for each of them.  We use \textit{Sent-512} to denote the average number of sentences within the first 512 tokens. Firstly, we sample the number of sentences within the same range as \textit{Sent-512} of each dataset, shown in Table 1 in supplementary materials.
In particular, we use a Poisson distribution $Poisson(\lambda)$ to sample sentence count for each article, where $\lambda$ equals to \textit{Sent-512} of each dataset.
Secondly, a gamma distribution $\Gamma(\alpha, \beta)$ is applied to sample the sentence length $l_i$ for each sentence sample, where $\alpha=2$ and $\beta=\bar{l}/2$. Finally, we sample scores for each sentence via a uniform distribution $U(0,1)$. We normalize the lengths on a set of target sizes of $\{75, 100, 125, 150, 175, 200, 225\}$ for CNewSum, $\{300, 350, 400, 450, 500, 550, 600\}$ for CNNDM, and normalize the sentence scores to sum up to 1. By now, we successfully simulate the profit (\textit{i.e.,} sentence score) and the size (\textit{i.e.,} sentence length) values to train a knapsack network.

\paragraph{Greedy method \& DP solver.} For the simulated data, we obtain the ground-truth indices of selected items using either dynamic programming (DP) as a solver on values, also named as profits~(\textit{i.e.,} sentence scores), or the greedy method as a solver on value densities (\textit{i.e.,} the scores divided by the length of sentences) for a sub-optimal solution. In another word, we use the DP solver or the greed method to produce ground-truth labels for training.
After labelling indices, 95\% of samples are used as a train set and the rest as a validation set. We use an 8-layer knapsack transformer with 8 heads and the model dimension of 768. 

\begin{table}[htbp]
  \centering{\footnotesize
  \resizebox{\linewidth}{!}{
    \begin{tabular}{cccc}
    \specialrule{0.15em}{3pt}{3pt}
    \textbf{Method} & \multicolumn{1}{p{3.5em}}{\textbf{Distribution}} & \multicolumn{1}{p{3em}}{\textbf{Error Rate} } & \multicolumn{1}{p{3em}}{\textbf{Matched Rate}} \\
    \midrule
    FPTAS-NN &  CNewSum & 2.7\% & 73.4\% \\
    Neural Knapsack & Pre-defined &      & 75.3\% \\
    Reinforcement Learning & Pre-defined &       & 92.7\% \\
    \midrule
    KT (DP) & CNewSum & 0.7\% & 93.8\% \\
    KT (DP) & CNNDM & 1.7\% & 82.1\% \\
    \midrule
    KT (Density Greedy)* & CNewSum      & 0.4\% & 96.8\% \\
    KT (Density Greedy)* & CNNDM  & 0.5\% & 93.2\% \\
    \specialrule{0.15em}{3pt}{3pt}
    \end{tabular}}%
        \caption{Comparison of Neural Knapsack Methods. The second column shows whether the data distribution is predefined with a fixed number of items, or sampled based on the distribution of the given dataset. KT denotes to our Knapsack Transformer. The asterisk mark denotes the method approximates sub-optimal solution given by the greedy algorithm.}
         \vspace{-1.0em}
  \label{tab:1}}%
\end{table}%

\begin{table*}[!htbp]
  \centering{\footnotesize
    \begin{tabular}{cccccccccc}
    \toprule
    \multirow{2}[2]{*}{\textbf{Method}} & \multicolumn{3}{c}{\textbf{CNNDM}} & \multicolumn{3}{c}{\textbf{NYT}} & \multicolumn{3}{c}{\textbf{CNewSum}} \\
          & \textbf{R1} & \textbf{R2} & \textbf{RL} & \textbf{R1} & \textbf{R2} & \textbf{RL} & \textbf{R1} & \textbf{R2} & \textbf{RL} \\
    \midrule
    Oracle & 54.70  & 30.40  & 50.80  & 61.90  & 41.70  & 58.30  & 46.84  & 30.54  & 40.08  \\
    Lead-3 & 40.50  & 17.70  & 36.70  & 35.50  & 17.20  & 32.00  & 30.43  & 17.26  & 25.33  \\
    TextRank (tf-idf) & 33.20  & 11.80  & 29.60  & 33.20  & 13.10  & 29.00  & 28.43  & 13.51  & 23.13  \\
    TextRank (BERT) & 30.80  & 9.60  & 27.40  & 29.70  & 9.00  & 25.30  & 29.35  & 15.56 & 24.17 \\
    TextRank (SimCSE) & 33.36  & 12.33  & 30.33  & 31.90     &  14.62    & 27.48    & 28.26  & 14.18  & 23.40  \\
    Adv-RF \citep{wang2018learning} & 35.51  & 9.38  & 20.98  & -     & -     & -     & -     & -     & - \\
    NeuSum \citep{zhou2018neural} & -     & -     & -     & -     & -     & -     & 30.61  & 17.36  & 25.66  \\
    TED \citep{wang2018learning}  & 38.73  & 16.84  & 35.40  & 37.78  & 17.63  & 34.33  & -     & -     & - \\
    PACSUM \citep{zheng2019sentence} & 40.70  & 17.80  & 36.90  & 41.40  & 21.70  & 37.50  & -     & -     & - \\
    STAS \citep{xu2020unsupervised} & 40.90  & \textbf{18.02}  & 37.21  & 41.46  & 21.80  & 37.57  & -     & -     & - \\
    FAR \citep{liang2021improving}  & 40.83  & 17.85  & 36.91  & \textbf{41.61}  & \textbf{21.88}  & 37.59  & -     & -     & - \\
    PMI \citep{padmakumar2021unsupervised}  & 36.60  & 14.52  & 23.30  & -     & -     & -     & -     & -     & - \\
    \midrule
    Ours (Ext-Doc) & 38.10  & 16.65  & 34.56  & 39.01  & 19.60  & 34.63  & 32.56  & 18.47  & 27.71  \\
    Ours (Rest-Ext) & 40.71  & 17.79  & 37.06  & 41.49  & 21.65  & 37.33  & 31.34  & 17.77  & 26.08  \\
    Ours (both) & \textbf{40.92}  & 17.88 & \textbf{37.27}  & 41.55  & 21.86  & \textbf{37.62}  & \textbf{32.82}  & \textbf{18.62}  & \textbf{27.84}  \\
    \bottomrule
    \end{tabular}%
    \caption{Comparison of unsupervised methods. For existing methods, the results of TextRank (SimCSE) is re-conducted by us, while other results are reported in existing papers \citep{zheng2019sentence, xu2020unsupervised, wang2021cnewsum}.}
    \vspace{-1.5em}
  \label{tab:2}}%
\end{table*}

\paragraph{Result.} As in the Table~\ref{tab:1},
We compare our \textit{Knapsack Transformer} to a set of existing neural knapsack methods including \textit{FPTAS-NN} \citep{hertrich2021provably}, \textit{Neural Knapsack} \citep{nomer2020neural}, and \textit{Reinforcement Learning} \citep{yildiz2022reinforcement}.
As the original \textit{FPTAS-NN} can not directly generate the optimal selection output, we apply an additional transformer to learn the mapping from its output vectors to the binary vectors of optimal selection in an end-to-end manner. For the other two alternatives, we list their reported results. 

Results show that our proposed \textit{Knapsack Transformer} (KT) outperforms baselines under the same data distribution, such as CNewSum.
In particular, the per item validation error (i.e., \textit{error rate} in the table header) of \textit{Transformer Knapsack} decreases to \textbf{0.7\%}, while the proportion of matching the optimal results (i.e., \textit{matched rate} in the table header) given by DP is about \textbf{93.8\%}. Our \textit{Transformer Knapsack} trained with greedy algorithm can even reach \textbf{96.8\%}, but note that this matched rate is compared to labels given by the greedy method, which is sub-optimal to knapsack problem.

Although the distributions of simulated data may differ in different work~\citep{hertrich2021provably, nomer2020neural, yildiz2022reinforcement},  we believe our \textit{Transformer Knapsack} has shown strong power as a trainable module to approximate the DP solver of knapsack problem compared to other choices. 


\subsection{Training of Joint Summarization Model}\label{Sec:4.2.2}

As in Figure~\ref{Fig:2}, for the encoder in extraction module and right-bottom siamese encoder, we apply the pretrained SimCSE model~\citep{gao2021simcse} from Huggingface.
We load the pre-trained knapsack transformer in our study. 
During the training, the parameters of knapsack transformer and encoder are not updated.
The scorer is a 2-layer transformer with 4 attention heads and 768 hidden size, and dimension of FFNN layer is 2048. For the transformer encoder and predictor in contrastive learning module, we use 4-layer transformers with 8 heads and 768 hidden size.
The seed is set to be 42. We apply two Adam optimizers with $\beta_1 = 0.9$, $\beta_2 = 0.999$ for training randomly initialized parameters and loaded pretrained parameters, respectively. We set the main learning rate $lr=3e-6$ for CNNDM and NYT and $lr=1e-6$ for CNewSum after a grid search from set \{3e-8, 1e-7, 3e-7, 1e-6, 3e-6, 1e-5\}, and the batch size is 64. We apply $\lambda_1=1$ and $\lambda_2=0.3$ for CNNDM/NYT and $\lambda_1=1$ and $\lambda_2=1$ for CNewSum as in Eq.~\eqref{Eq:final_loss}. We compare the performance of a set of candidate models for summarization with the target length limits of 100, 150, 200 characters for CNewSum, and 400, 450, 500 letters for CNNDM.



\subsubsection{Baselines.} \label{Sec:4.3}

We compare our method with: (a) Lead-based algorithm; (b) Textrank with different types of sentence representations \citep{erkan2004lexrank}; These methods provide explicit sentence scores that can be used by DP. We also compare the proposed models with a set of existing models in the case without length control, including Adv-RF \citep{wang2018learning}, TED \citep{yang2020ted}, PACSUM \citep{zheng2019sentence}, STAS \citep{xu2020unsupervised}, PMI \citep{padmakumar2021unsupervised}, NeuSum \citep{zhou2018neural}, etc. 

\subsubsection{Automatic Evaluation.}\label{Sec:4.4}

For automatic evaluation, model checkpoints are saved and evaluated on the validation set every 1,000 steps. The top-3 checkpoints are then selected based on the validation loss, and the averaged results of checkpoints on the test set are reported.


\begin{table}[htbp]
  \centering{\footnotesize
    \begin{tabular}{C{1.9cm}C{0.25cm}C{0.25cm}C{0.25cm}C{0.25cm}C{0.25cm}C{0.25cm}C{0.25cm}C{0.25cm}C{0.25cm}} 
    \toprule
    \multirow{2}[2]{*}{\textbf{Method}} & \multicolumn{3}{c}{\textbf{CNNDM}} & \multicolumn{3}{c}{\textbf{NYT}} & \multicolumn{3}{c}{\textbf{CNewSum}} \\
          & \textbf{R1}    & \textbf{R2}    & \textbf{RL}    & \textbf{R1}    & \textbf{R2}    & \textbf{RL}  & \textbf{R1}    & \textbf{R2}    & \textbf{RL} \\
    \midrule
    Default & 40.9  & 17.9  & 37.3  & 41.6  & 21.9  & 37.6 & 32.8  & 18.6  & 27.8 \\
    w/o Ext-Doc & 40.7  & 17.8  & 37.1  & 41.5  & 21.7  & 37.3 & 31.3  & 17.8  & 26.1  \\
    w/o Rest-Ext & 38.1  & 16.6  & 34.6  & 39.0  & 19.6  & 34.6 & 32.6  & 18.5  & 27.7 \\
    w/o Predictor & 26.2  & 11.5  & 23.4  & 33.6  & 14.8  & 29.1 & 30.4  & 15.4  & 25.3  \\
    w/o Sent. Enc. & 37.9  & 16.5  & 34.2  & 39.6  & 20.0  & 35.2 & 30.6  & 15.5  & 25.5  \\
    w/o Neg. Sam. & 38.5  & 16.8  & 35.3  & 35.1  & 16.4  & 30.8 & 31.2  & 16.2  & 36.1 \\
    \bottomrule
    \end{tabular}
    \caption{Ablation study of on CNNDM, NYT and CNewSum.}
    \vspace{-1em}
  \label{tab:ablation}}%
\end{table}%


The automatic evaluation is reported in Table~\ref{tab:2}. In general, the performance of our proposed method is very competitive with strong state-of-the-art models for unsupervised summarization. On all three datasets, our model using both two objectives largely outperform the most related baseline by using graph-based TextRank with SimCSE. We also notice that for CNNDM and NYT, the Rest-Ext objective alone can already bring quite good performance. It evidences the effectiveness of our proposed prediction objectives.

An ablation study the three datasets is given in Table~\ref{tab:ablation}, which shows the effect of different learning objectives and model components. Although the training flow can work without the predictor and second-level encoder, results show that they contribute significantly to the performance. Without negative sampling, it is likely that learning will collapse to trivial states as the selector can not be well-trained as discussed in Method section. 

\subsubsection{Length Control.}
Results with the length control for CNNDM and CNewSum are provided in Table~\ref{tab:control_cnndm} and Table~\ref{tab:control_cnewsum}, respectively, where ``TR'' denotes TextRank, ``KTDP'' and ``KTDEN'' denote our Knapsack Transformer trained to approximate DP solver and greedy density solver, respectively. In the inference, we apply dynamic programming (DP) for 0-1 knapsack problem on the pre-trained scorers using different methods, which means that the final selecting of sentences is not solely based on score ranking thus probably with a performance drop. We can see by using the end-to-end training with our Knapsack Transformer, the performance of the scorer is consistently increased in a variety of target lengths, in terms of both automatic metrics and the distance to target lengths. Results show that our proposed length control method brings more potentials towards length-controllable summarization.

\begin{table}[htbp]
  \centering{\footnotesize
    \begin{tabular}{C{0.5cm}C{2.8cm}C{0.4cm}C{0.4cm}C{0.4cm}C{1.2cm}} 
    \toprule
    \textbf{Target} & \textbf{Method} & \textbf{R1}    & \textbf{R2}    & \textbf{RL}    & \multicolumn{1}{c}{\textbf{Ave. Len}} \\
    \midrule
    \multirow{5}[2]{*}{400} & TR (tf-idf)+ DP & 31.7  & 10.5  & 28.8  & 392 (13) \\
          & TR (SimCSE)+ DP & 30.4  & 9.6   & 27.5  & 381 (20) \\
          & Ours + DP & 38.6  & 16.9  & 35.0  & 380 (19) \\
          & Ours + TFDP + DP & 39.0  & \textbf{17.1}  & \textbf{35.4}  & 381 (18) \\
          & Ours + TFDEN + DP & \textbf{39.2}  & \textbf{17.1}  & 35.5  & 380 (9) \\
    \midrule
    \multirow{5}[2]{*}{450} & TR (tf-idf)+DP & 32.3  & 10.9  & 29.5  & 443 (15) \\
          & TR (SimCSE)+ DP & 31.4  & 10.4  & 28.5  & 428 (21) \\
          & Ours + DP & 39.5  & 17.3  & 35.9  & 430 (19) \\
          & Ours + KTDP + DP & \textbf{40.0}  & \textbf{17.5}  & \textbf{36.4}  & 433 (17) \\
          & Ours + KTDEN + DP & 39.2  & 17.1  & 35.8  & 431 (11) \\
    \midrule
    \multirow{5}[2]{*}{500} & TR (tf-idf)+DP & 32.5  & 11.1  & 29.8  & 494 (13) \\
          & TR (SimCSE)+ DP & 31.7  & 10.5  & 28.7  & 480 (23) \\
          & Ours + DP & 39.8  & 17.4  & 36.3  & 480 (20) \\
          & Ours + KTDP + DP & \textbf{40.5}  & \textbf{17.7}  & \textbf{37.0}  & 481 (18) \\
          & Ours + KTDEN + DP & 40.4  & \textbf{17.7}  & \textbf{37.0}  & 480 (11) \\
    \bottomrule
    \end{tabular}%
    \caption{Comparison of Length control mehods on CNNDM. We provide the average and standard deviation (in parentheses) of the summary lengths in the last column.}
    \vspace{-1em}
  \label{tab:control_cnndm}}%
\end{table}%

\begin{table}[htbp]
  \centering{\footnotesize
    \begin{tabular}{C{0.5cm}C{2.8cm}C{0.4cm}C{0.4cm}C{0.4cm}C{1.2cm}} 
    \toprule
    \textbf{Target} & \textbf{Method} & \textbf{R1}    & \textbf{R2}    & \textbf{RL}    & \multicolumn{1}{c}{\textbf{Ave. Len}} \\
    \midrule
    \multirow{5}[2]{*}{100} & TR(tf-idf)+DP & 28.0  & 13.2  & 22.3  & 89 (17) \\
          & TR(SimCSE)+DP & 28.1  & 13.6  & 22.2  & 88 (18) \\
          & Ours+DP & 33.0  & 18.7  & 27.7  & 89 (15) \\
          & Ours+KTDP+DP & \textbf{33.5}  & \textbf{19.0}  & \textbf{28.2}  & 89 (12) \\
          & Ours+KTDEN+DP & 33.2  & 18.8  & 28.0  & 88 (15) \\
    \midrule
    \multirow{5}[2]{*}{150} & TR(tf-idf)+DP & 28.5  & 13.5  & 22.8  & 136 (25) \\
          & TR(SimCSE)+ DP & 28.6  & 14.3  & 23.3  & 135 (25) \\
          & Ours+DP & 32.7  & 18.6  & 27.5  & 136 (22) \\
          & Ours+KTDP+DP & \textbf{33.1}  & \textbf{18.8}  & \textbf{28.0}  & 136 (23) \\
          & Ours+KTEN+DP & \textbf{33.1}  & 18.7  & 27.9  & 136 (21) \\
    \midrule
    \multirow{5}[2]{*}{200} & TR(tf-idf)+DP & 28.1  & 13.5  & 22.9  & 180 (35) \\
          & TR (SimCSE)+DP & 28.4  & 14.1  & 23.3  & 180 (34) \\
          & Ours+DP & 30.6  & 17.4  & 25.7  & 181 (31) \\
          & Ours+KTDP+DP & 30.9  & 17.5  & 25.9  & 180 (30) \\
          & Ours+KTDEN+DP & \textbf{31.9}  & \textbf{18.1}  & \textbf{27.2}  & 181 (30) \\
    \bottomrule
    \end{tabular}%
    \caption{Length control methods on CNewSum.}
    \vspace{-1em}
  \label{tab:control_cnewsum}}%
\end{table}%

\subsubsection{Human Evaluation.}\label{Sec:5}

We further conduct a manual evaluation on the CNNDM dataset with a target length of 500. We randomly sample 1000 entries from the test set and generate summaries with three candidate models. Each sample is scored between 1 and 5 points by three participants from Amazon Mechanical Turk to answer three questions:
(a) Whether the summaries capture the key points of the source document (Relevance or Rele.); (b) the sentence-level coherence of the summaries (Coherence or Cohe.); (c) the factual consistency of the summaries (Consistency or Cons.). Results in Table~\ref{tab:human_eval} show that our proposed unsupervised model using knapsack transformer achieves the best score in terms of relevance and consistency. 
Meanwhile, we find that our unsupervised models perform even better than the supervised Bertsum model \citep{liu2019fine} according to judge scores. It indicates that rouge scores might not capture the real quality of the extracted summaries, while our proposed unsupervised learning methods could outperform supervised ones from the aspect of human readers. Overall, the human evaluation further confirms the effectiveness of our method.
\begin{table}[htbp]
  \centering{\footnotesize
    \begin{tabular}{cccc}
    \toprule
    \textbf{Model} & \textbf{Rele.} & \textbf{Cohe.} & \textbf{Cons.} \\
    \midrule
    \tabincell{l}{Bertsum + DP} & 3.268 & 3.248 & 3.306 \\
    \tabincell{l}{Ours + DP} & 3.293 & \textbf{3.256} & 3.347\\
    \tabincell{l}{Ours + KTDP + DP} & \textbf{3.329}& 3.217 & \textbf{3.393}\\
    \bottomrule
    \end{tabular}}%
    \caption{Human evaluation of different model outputs on CNNDM. We provide the average scores judged by human annotators. The improvements for relevance and consistency are significant by two-sample statistical tests (p<0.05).}
     \vspace{-1em}
  \label{tab:human_eval}
\end{table}%


\section{Conclusion}\label{Sec:7}

In this paper, we introduce a novel unsupervised extractive summarization model based on siamese network and a bidirectional prediction objective. It does not rely on position-related assumption and can be applied for end-to-end training. Meanwhile, by introducing the knapsack transformer that approximates the DP solver of knapsack problem, the length control strategy can also be trained in an end-to-end manner. Experiments demonstrate that our proposed unsupervised method largely outperforms related centrality-based baselines and comparable with the state-of-the-art unsupervised methods. For the length control, our method outperforms baselines in a set of target lengths on both English and Chinese datasets, and the use of end-to-end length control training consistently provides a performance gain, making another step towards length-controllable summarization. Moreover, our method achieves higher average rating scores in human evaluation even compared with supervised BERTsum. We believe the model performance can be further improved by using better sentence and document representations.

In general, supervised methods largely rely on the quality of annotated data, while this is quite tricky in summarization tasks where more tedious efforts are required for annotators to read and understand articles. Poorly annotated summaries limit supervised methods, but unsupervised methods can always take advantage of the up-to-date and large-scale data from various sources. Therefore, we believe that unsupervised methods have the potential for more wide applications, which are still desired to be further explored.

\newpage
\bibliography{Reference}

\newpage

\appendix
\section{Appendix} 
\subsection{Motivation of Length-Controlled Adaptation} \label{diagram_lenth_control}

\begin{figure}[htbp]
\begin{center}
 \includegraphics[width=1.0\linewidth]{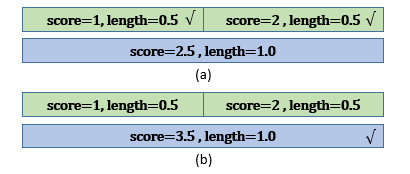}
 \caption{The diagram showing the shortcoming of directly using score-based length planning without adaptation. It shows the lengths and scores of 3 sentences. Both (a) and (b) ensure an order that $s_3>s_2>s_1$ and have the same topK selections without length constraints, but a small change of $s_3$ could result in totally different selection with the maximum length as 1. Therefore, the scores need to be accurately scaled to the planning algorithm and thus provide a selection with better performance.} \label{Fig:trade_off}
\end{center}
\end{figure}

Although the sentences scores given by extractive summarization methods such as Bertsum can already serve as the profits for using knapsack algorithm to control lengths, they are not necessarily in the optimal scale as the model is initially trained without the planning process of knapsack. We show one example in Figure~\ref{Fig:trade_off}. To make the optimal selection according to the goodness of sentence representations under length constraints, it is important to \textbf{make the scaling of scores adaptive to the planning algorithm}. We achieve it by training the scorer jointly with the differentiable knapsack transformer, and we freeze the parameters of knapsack transformer since it has been pre-trained as a knapsack solver. Meanwhile, as it is tedious to annotate ground-truth summaries under different length control, we need to develop an unsupervised architecture for end-to-end training.

\subsection{Visualizing and Further Explanation of Bidirectional Prediction Objective}

\begin{figure}[!tbp]
\begin{center}
 \includegraphics[width=1.0\linewidth]{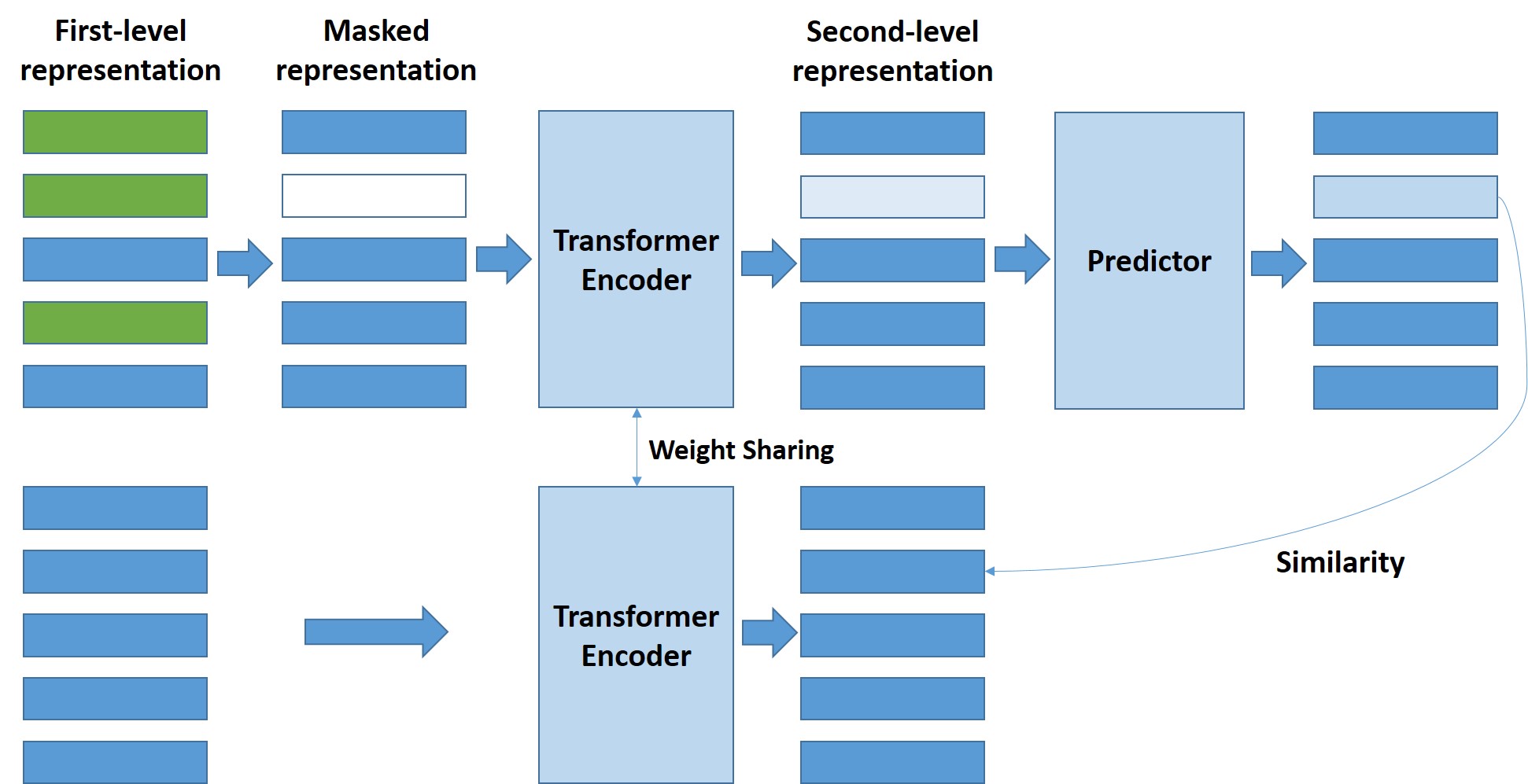}
 \caption{The diagram of visualizing \textbf{Rest-Ext} objective. We masked out a randomly sampled sentence from the selected set, and use all the other sentences (no matter selected or unselected) to predict it. Compared with unselected sentences, selected sentences from an extractive summary should be better predicted by the remaining sentences. Thus, this learning objective enables our scorer to select sentences with a higher chance to be accurately predicted.} \label{Fig:Rest-Ext}
\end{center}
\end{figure}

\begin{figure}[!tbp]
\begin{center}
 \includegraphics[width=1.0\linewidth]{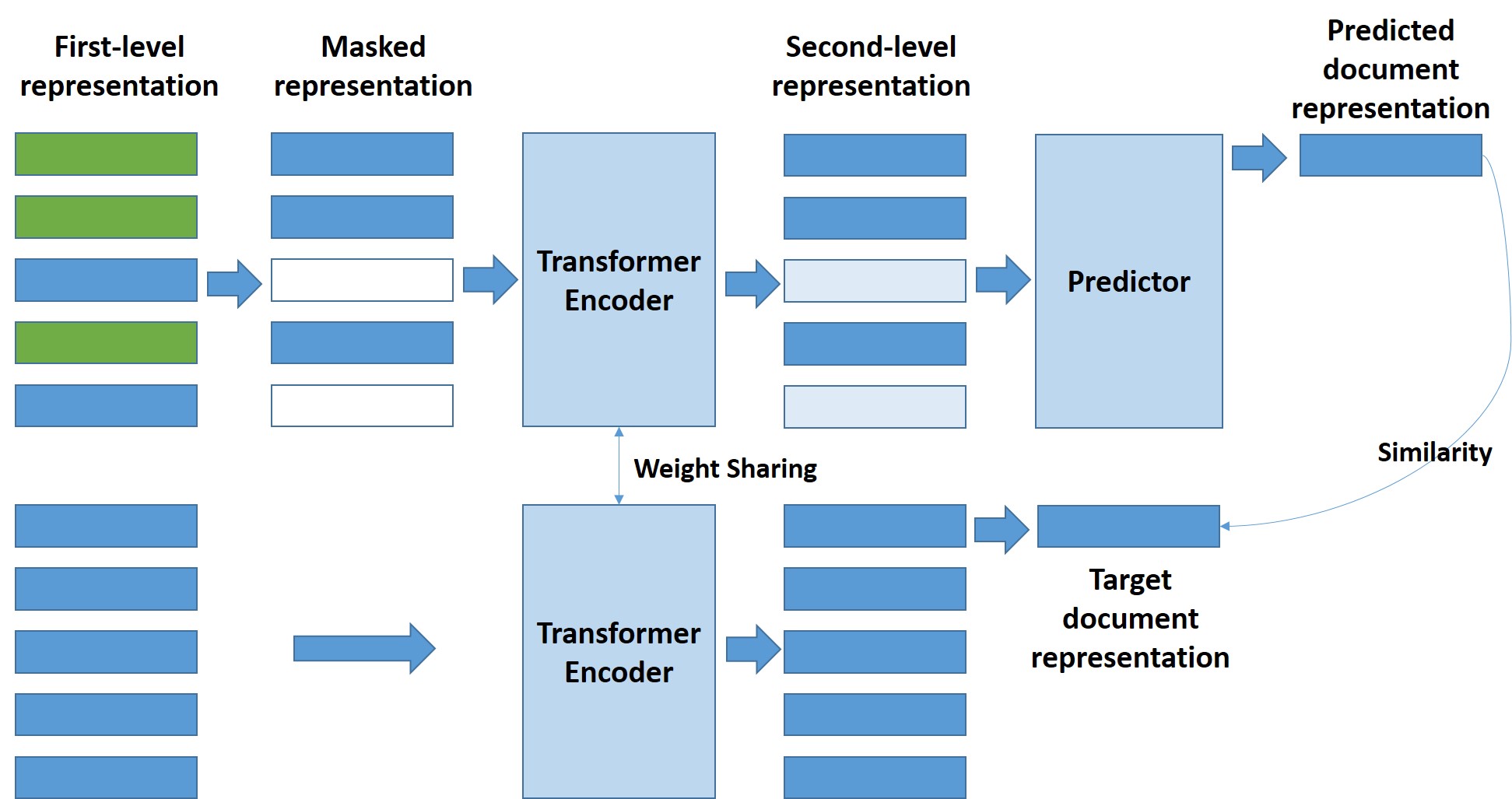}
 \caption{The diagram of visualizing \textbf{Ext-Doc} objective. We masked out all the unselected sentences, and only use the set of selected sentences to predict the whole document representation. Thus, this learning objective enables our scorer to select the sentences that are more likely to predict or represent the whole document.} \label{Fig:Ext-Doc}
\end{center}
\end{figure}

In this paper, we introduced a bidirectional prediction objective, where we mainly explained it with the mathematical formula. To make it more clear, we further provide detailed diagrams for \textbf{Rest-Ext} objective in Figure~\ref{Fig:Rest-Ext}, and \textbf{Ext-Doc} objective in Figure~\ref{Fig:Ext-Doc}. The first-level representation is the sentence representation (\texttt{[CLS]}) given by encoders such as BERT and SimCSE. The green bars are the selected sentences based on Gumble topK (without knapsack transformer) or rounding (with knapsack transformer) for building the extractive summarization, where the selecting process is differentiable as is discussed in the main paper. The white bars are the masked representation with zero vectors. 

Note that in these two figures, we only show the positive examples. In terms of the negative examples, the green bars should turn to be a subset of unselected sentences with the lowest scores, and the similarities should be minimized rather than maximized. Moreover, only the gradients from maximizing the positive similarities are applied in updating the predictor, although all the gradients pass the predictor and are applied to update the scorer and transformer encoder. This is because minimizing the negative similarity can harm the predictor's performance when its gradients update the predictor. In implementing this mechanism, we apply another weight copy and stop-gradient way like that is used in a siamese network.

After training, the scorer is able to give higher scores for sentences that are \textbf{more likely to be predicted by the remaining sentences or more likely to predict the whole document representation}. Thus, in the inference, by selecting the sentences with higher scores, we are selecting the sentences with high centrality based on bidirectional prediction criterion. 

\subsection{Datasets}\label{appendix_datasets}

\paragraph{CNNDM.} The CNN/DailyMail dataset \citep{hermann2015teaching} contains news articles from the CNN and Daily Mail websites, with labelled abstractive and extractive summaries. There are 287,226 training samples, 13,368 validation samples and 11,490 test samples. 
\paragraph{CNewSum.} It is a Chinese news article dataset with labelled abstractive and extractive summaries \citep{wang2021cnewsum}. It contains 304,307 samples including 275,596 training samples, 14,356 validation samples and 14,355 test samples. 

\paragraph{NYT.} The NYT dataset ~\citep{durrett2016learning} contains 110,540 article with abstractive summary collected from New York Times newspaper. We follow its paper to split the original dataset into 100,834 training and 9,706 test examples.

\paragraph{Table~\ref{tab:datasets}} shows statistics of CNNDM and CNewSum datasets regarding the distribution of sentence length and sentence number, which is used to generate the simulated data in training knapsack transformer. The \textit{Sent-512} denotes the average number of sentences within the first 512 tokens. Note that we use the string length as the criteria of lengths.


\vspace{-0.5em}
\begin{table}[htbp]
  \centering{\footnotesize
    \begin{tabular}{p{7.69em}cccc}
    \specialrule{0.15em}{3pt}{3pt}
    \multicolumn{1}{c}{\multirow{2}[4]{*}{}} & \multicolumn{2}{c}{\textbf{CNewSum}} & \multicolumn{2}{c}{CNNDM} \\
\cmidrule{2-5}    \multicolumn{1}{c}{} & \textbf{Mean}  & \textbf{St.dev} & \textbf{Mean}  & \textbf{St.dev} \\
    \midrule
    \multicolumn{1}{c}{Article Length} & 1,175 & 1,109 & 3,893 & 2,001 \\
    \multicolumn{1}{c}{Sentence Length} & 69.7  & 61.7  & 121.1 & 55.9 \\
    \multicolumn{1}{c}{\# Sent. per Art.} & 16.8  & 17.5  & 32.1  & 17.9 \\
    \multicolumn{1}{c}{\# Sent-512} & 9.3  & -  & 17.4  & - \\
    \specialrule{0.15em}{3pt}{3pt}
    \end{tabular}%
    \caption{Statistics of datasets}
  \label{tab:datasets}}%
\end{table}%

\subsection{Visualization of Learning Curves}

\begin{figure*}[!tbp]
\begin{center}
 \includegraphics[width=1.0\linewidth]{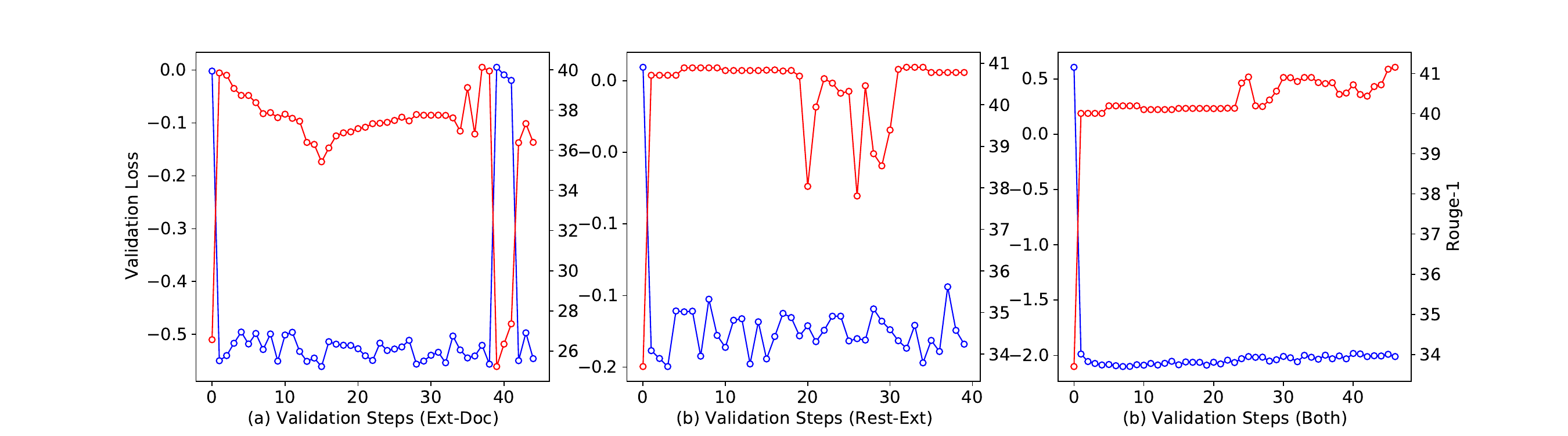}
 \caption{The learning curves of unsupervised training with three different objectives on CNNDM. (a) Ext-Doc objective only. (b) Rest-Ext objective only. (c) Using both two objectives. The red curves are Rouge-1 w.r.t. the validation steps, and the blue curves are validation loss w.r.t. the validation steps.} \label{Fig:Learning_Curves}
\end{center}
\end{figure*}

The learning curves of different objectives are given in Figure~\ref{Fig:Learning_Curves}. We provide the curves of validation loss and Rouge-1 for training on CNNDM in each case. We perform a validation for every 1000 training iterations as a validation step. In general, the convergent speed is generally fast for our proposed model. However, for the case (a) with \textbf{Ext-Doc} objective only, the learning curve is not perfectly stable, which also explains why its performance is not as good as \textbf{Rest-Ext.} Overall, using \textbf{both two objectives} makes the training process perfectly stable, which is quite consistent to our main results in the paper.

Note that the training objective of an effective unsupervised learning method should well align with evaluation metrics. Thus, a decrease of validation loss is expected to be associated with an increase of rouge scores during the training, which is evidenced by our case (3) in Figure~\ref{Fig:Learning_Curves}. This visualization of learning curves further confirms the effectiveness of our proposed learning objectives.
Therefore, in our paper, we can easily select the checkpoints based on the validation loss rather than validated rouge scores. 

\end{document}